\documentclass[letterpaper, 10 pt, conference]{ieeeconf}

\IEEEoverridecommandlockouts

\title{
  LOTUSim-Energy: A Maritime Simulator for Human-Drone Interaction in Autonomous Offshore Operation \& Maintenance
}

\PassOptionsToPackage{table}{xcolor}
\usepackage{graphicx}
\usepackage{tabularx}
\usepackage{multirow}
\usepackage{amsmath, algorithmic, algorithm}
\usepackage{dingbat}

\usepackage{graphicx}
\usepackage{booktabs}
\usepackage{array}
\usepackage{tabularx}
\usepackage{hyperref}
\hypersetup{hidelinks}
\usepackage{footnote}
\usepackage{pifont}
\usepackage{multirow}
\usepackage{caption}
\usepackage{tikz}
\usepackage{framed}
\usepackage{arydshln}

\usepackage{svg}
\usepackage{makecell}
\usepackage{subcaption}

\usepackage{xcolor}
\definecolor{lightblue}{RGB}{220,230,255}
\usepackage{tcolorbox}
\tcbuselibrary{listingsutf8}

\let\labelindent\relax
\usepackage{enumitem,amssymb}
\newlist{todolist}{itemize}{2}
\setlist[todolist]{label=$\square$}
\usepackage{pifont}

\begin{document}

\author{
    Juliette Grosset$^{1}$,
    Marie Dubromel$^{1}$,
    H\'el\`ene Lech\^ene$^{1}$,
    Quentin Arzel$^{1}$,
    C\'edric Buche$^{2}$
    \\[2pt]
    \footnotesize
    $^{1}$CROSSING IRL 2010, Naval Group \quad
    $^{2}$CROSSING IRL 2010, CNRS, IMT Atlantique
}

\maketitle

\begin{abstract}

Offshore maintenance requires operations in the air, the surface, and the subsea domain and include human supervision.
This paper presents \emph{LOTUSim-Energy}, a real-time maritime simulator designed for multi-domain human–drone interaction for offshore operation and maintenance.
The platform unifies heterogeneous unmanned vehicles (Unmanned Aerial Vehicles: UAVs, Unmanned Surface Vehicles: USVs, Autonomous Underwater Vehicles: AUVs, Remotely Operated Vehicles: ROVs) within a distributed architecture coupling environmental forcing (wind, waves, currents) and provides immersive user interfaces for supervision (desktop and virtual reality).
A structured offshore task library enables repeatable evaluation of autonomy stacks under realistic metocean disturbances. The simulator supports realistic physics, energy-aware battery modeling, and fault-detection pipelines as modular validation tools.
System-level performance is demonstrated on a multi-domain inspection scenario for monopile and transition piece structure, where we evaluate the reliability of integrated waypoint-follower plugin and Automatic Identification System (AIS)-referenced trajectory tracking under real-time energy monitoring.
By combining unified environmental physics, heterogeneous vehicle simulation, and immersive supervision, LOTUSim-Energy provides an integration testbed for prototyping and rehearsing offshore human–robot collaboration workflows, as a step toward de-risking sea deployment.

\end{abstract}

\begin{keywords}
Marine robotics, Simulator, Real-time Maritime Simulation, Offshore Operation \& Maintenance, Energy, Ocean Current Modelling, Human-In-The-Loop.
\end{keywords}

\section{Introduction}

The maintenance of offshore wind farms is shifting toward human-supervised multi-robot teams. However, deploying such teams remains a formidable challenge: aerial, surface, and subsea assets must coordinate while battling tightly coupled wind, wave, and current disturbances. In these environments, a single perception failure or a miscalculation in energy reserves can lead to catastrophic asset loss. Given the narrow weather windows and stringent safety constraints, high-fidelity simulation is a non-negotiable prerequisite for de-risking human-robot coordination and validating integrated autonomy stacks before sea trials.

To safely deploy autonomous fleets in harsh maritime environments, we require a simulation framework capable of integrating two interdependent pillars:

\begin{enumerate}
    \item Realistic Environmental Physics: ocean currents, waves and aerial disturbance (wind).
    \item Human-in-the-Loop (HITL): interactive supervision affects the reliability of multi-robot coordination.
\end{enumerate}

Existing simulation platforms \cite{Song2025OceanSim, romrell2025previewholoocean20, marineGym, Grimaldi2025}, excel in one domain, from high-fidelity rendering to sensors or AI training, but fail to integrate HITL and environmental physics.

The primary contribution of this paper is \emph{LOTUSim-Energy}\footnote{LOTUSim-Energy: \url{https://www.youtube.com/watch?v=NRd-LjRrrac}}, an integrated simulation ecosystem for maritime robotics providing:

\begin{itemize}
    \item \textit{An Interactive Multi-Domain Simulator}: A unified architecture coupling cross-domain vehicle dynamics (air/surface/subsea) with physics-consistent metocean fields (wind, waves, and currents). It provides a modular, multi-user interface for immersive supervision, manual takeover, and mission procedure rehearsal, built on \emph{LOTUSim} \cite{LOTUSim26iros}.

    \item \textit{A Benchmarking Task Library}: A configurable set of offshore Operation \& Maintenance (O\&M) scenarios for multi-domain inspection. This library offers repeatable evaluation of heterogeneous autonomy systems and HITL intervention under realistic maritime disturbances.

    \item \textit{An Integrated Offshore Autonomy Stack}: A modular suite designed for heterogeneous coordination and mission-level monitoring, including:
    \begin{itemize}
        \item Navigation \& Tracking: AIS-referenced trajectory followers and geometric waypoint controllers for UAV, USV, and subsea assets.
        \item Mission-Aware Sensing: A battery simulation plugin for real-time charge/discharge monitoring based on propulsion effort, and a YOLO-based vision pipeline for detecting structural cracks and corrosion.
    \end{itemize}

\end{itemize}

\section{Context}
\label{sec:context}
This section summarises offshore wind technology and economics, emphasising the O\&M challenges that motivate robotics and high-fidelity simulation.

\subsection{Offshore Wind Turbine Technology \& O\&M}
\label{subsec:wt_techno}
Modern offshore wind turbines convert kinetic energy in the wind into electrical power. Power is exported to shore through array and export subsea cables and one or more offshore substations.
The lifecycle of offshore wind turbines comprises four phases\footnote{Lifecycle of offshore wind: \url{https://tinyurl.com/2s39zh7p}}: planning \& development, installation \& construction, O\&M (training, logistics, inspection, servicing, and balance-of-plant maintenance), and decommissioning.

Long-horizon O\&M remains technically challenging and costly due to harsh marine conditions, limited weather windows, and complex operational coordination. Robotics and digitalisation are expected to shift O\&M from manual, reactive practices toward remote, data-driven, and proactive operations. Aerial and underwater vehicles enable safer inspections, faster survey cycles (e.g., blade inspection times projected to decrease by up to $\sim$40\%\footnote{\url{https://tinyurl.com/529tuts7}}), and improved predictive monitoring. Realising these benefits requires planning, control, and validation tools capable of evaluating strategies under realistic environmental and operational constraints, motivating task- and system-level simulators.

O\&M typically contributes \mbox{15–35\%} of lifetime costs\footnote{\url{https://tinyurl.com/3u4x6uuw}}.

\subsection{Simulation for Offshore Wind O\&M requirements}

Autonomous O\&M of offshore wind farms involves coordinated deployment for heterogeneous robotic systems across multiple domains, including aerial (Unmanned Aerial Vehicle, UAV), surface (Unmanned Surface Vehicle, USV), and subsea (Autonomous Underwater Vehicle, AUV / Remotely Operated Vehicle, ROV) platforms. Therefore, reliable deployment of such systems requires realistic simulation environments supporting cross-domain physics for drone swarms. In addition, human interaction is essential for supervision and mission validation.
A simulation platform suitable for offshore O\&M must therefore provide:
\begin{enumerate}
    \item realistic environmental physics (air--surface--subsea);
    \item Human-in-the-Loop (HITL) interaction
\end{enumerate}

\subsection{Multi-Domain Simulation for Maritime}

Several platforms \cite{Song2025OceanSim, romrell2025previewholoocean20, marineGym, Grimaldi2025} offer high-fidelity rendering but lack native support for HITL interaction. Such capabilities are essential for offshore operations, where human supervision and intervention play a critical role.

Many simulators also fall short in realistic environmental physics, particularly in modeling underwater current disturbances. Common approaches rely on static or user-defined flow fields, including precomputed fixed CFD environments (USVsim \cite{Paravisi2019}), constant and unidirectional currents (LRAUVSim \cite{Player2023MultiAUV} and MarineGym \cite{marineGym}), or simplified profiles (Stonefish \cite{Grimaldi2025} and HoloOcean \cite{romrell2025previewholoocean20}). Such simplifications limit the study of realistic offshore conditions, where the combined effects of air, surface, and underwater disturbances can significantly influence vehicle behavior.

Some simulators focus specifically on offshore wind applications, but these are primarily designed for logistics planning, asset management, or cost analysis rather than robotics-centric mission rehearsal. While recent efforts (e.g., Northwind \cite{stadtmann2023northwind}) demonstrate growing interest in autonomous O\&M, comprehensive reviews \cite{aldhaheri2025underwaterroboticsimulatorsreview} highlight the persistent lack of unified environments that integrate HITL interaction, realistic underwater physics, and multi-domain operations. Representative maritime robotics simulators, spanning both offshore-wind-specific and general-purpose platforms, are compared in Table~\ref{tab:energy-simulators}; \emph{LOTUSim-Energy} leverages key features of existing platforms (e.g. wind turbine model, O\&M planning) while introducing several novel contributions:
\begin{itemize}
    \item Unified air, surface, and subsea physics with dynamic
environmental forces;
    \item A current model enabling depth-dependent drift and
shear effects;
    \item Immersive HITL interfaces for real-time and multi-
operator mission rehearsal
\end{itemize}

By combining high-fidelity environmental modeling with advanced human–robot interaction capabilities, \emph{LOTUSim-Energy} enables realistic validation of autonomous O\&M strategies prior to deployment.

\begin{table*}[htbp!]
    \centering
    \caption{State-of-the-art simulator platforms for autonomous O\&M and immersive operator decision-making. N/A: not reported by the cited work.}
    \label{tab:energy-simulators}
    \scriptsize
    \resizebox{\textwidth}{!}{%
    \begin{tabular}{|l|c|c|c|c|c|c|c|c|c|}
    \hline
    & \textbf{MIMRee} \cite{jovan2021minree}
    & \textbf{Northwind} \cite{stadtmann2023northwind}
    & \textbf{ACOMAR} \cite{pederson2019acomar}
    & \textbf{ROMEO} \cite{romeo2020report}
    & \textbf{OceanSim} \cite{Song2025OceanSim}
    & \textbf{HoloOcean} \cite{romrell2025previewholoocean20}
    & \textbf{MarineGym} \cite{marineGym}
    & \textbf{Stonefish} \cite{Grimaldi2025}
    & \textcolor{teal}{\textbf{LOTUSim-Energy}} \\
    \hline

    Country
    & UK
    & Norway
    & Denmark
    & EU
    & N/A
    & N/A
    & N/A
    & N/A
    & France/Autralia \\
    \hline

    Domain
    & Surface
    & Air
    & Underwater
    & Air
    & Underwater
    & Underwater
    & Underwater
    & \makecell{Underwater + Surface\\+ Ground}
    & Air + Surface + Underwater \\
    \hline

    Drone Integration
    & \checkmark
    & \textcolor{red}{$\times$}
    & \checkmark
    & \textcolor{red}{$\times$}
    & \checkmark
    & \checkmark
    & \checkmark
    & \checkmark
    & \checkmark \\
    \hline

    Rendering Engine
    & Gazebo
    & Unity
    & Gazebo
    & Custom
    & Isaac Sim
    & Unreal Engine
    & Isaac Sim
    & Custom (OpenGL)
    & Unity \\
    \hline

    VR / AR
    & \textcolor{red}{$\times$}
    & \checkmark
    & \textcolor{red}{$\times$}
    & \checkmark
    & \textcolor{red}{$\times$}
    & \textcolor{red}{$\times$}
    & \textcolor{red}{$\times$}
    & \textcolor{red}{$\times$}
    & \checkmark + Multi-user \\
    \hline

    Energy Model
    & Statistical
    & \textcolor{red}{$\times$}
    & \textcolor{red}{$\times$}
    & \textcolor{red}{$\times$}
    & N/A
    & N/A
    & N/A
    & N/A
    & Theoretical methods \\
    \hline

    Data Fusion
    & \textcolor{red}{$\times$}
    & \textcolor{red}{$\times$}
    & \checkmark
    & \textcolor{red}{$\times$}
    & N/A
    & N/A
    & N/A
    & N/A
    & \textcolor{red}{$\times$} (On-Going)  \\
    \hline

    Wind Turbine Model
    & \textcolor{red}{$\times$}
    & \checkmark
    & \textcolor{red}{$\times$}
    & \checkmark
    & \textcolor{red}{$\times$}
    & \textcolor{red}{$\times$}
    & \textcolor{red}{$\times$}
    & \textcolor{red}{$\times$}
    & \checkmark \\
    \hline

    O\&M Planning
    & \checkmark
    & \checkmark
    & \textcolor{red}{$\times$}
    & \checkmark
    & \textcolor{red}{$\times$}
    & \textcolor{red}{$\times$}
    & \textcolor{red}{$\times$}
    & \textcolor{red}{$\times$}
    & \checkmark \\
    \hline

    Weather Prediction
    & \textcolor{red}{$\times$}
    & \checkmark
    & \textcolor{red}{$\times$}
    & \checkmark
    & \textcolor{red}{$\times$}
    & \textcolor{red}{$\times$}
    & \textcolor{red}{$\times$}
    & \textcolor{red}{$\times$}
    & \textcolor{red}{$\times$} (Real-time) \\
    \hline
    \end{tabular}%
    }
\end{table*}

\section{LOTUSim-Energy Architecture}
\label{sec:LOTUSim}

\emph{LOTUSim-Energy} is a distributed, server–client simulation framework designed for multi-domain robotics. {Gazebo} serves as the central orchestrator for asset management and simulation timing (deterministic step scheduler), while separate client modules execute specific simulation tasks. The core simulation control module interfaces with three primary client modules:

\begin{itemize}
    \item Physics: \emph{LOTUSim-Xdyn} as the physics interface,
    \item Agent Interaction: \emph{ROS~2} for inter-agent messaging and hardware-in-the-loop bridges,
    \item Rendering: \emph{Unity} as the optional high-fidelity renderer for human–robot interaction (HRI).
\end{itemize}

\begin{figure}[htbp]
    \centering
    \captionsetup{justification=centering}
    \includegraphics[width=0.50\linewidth]{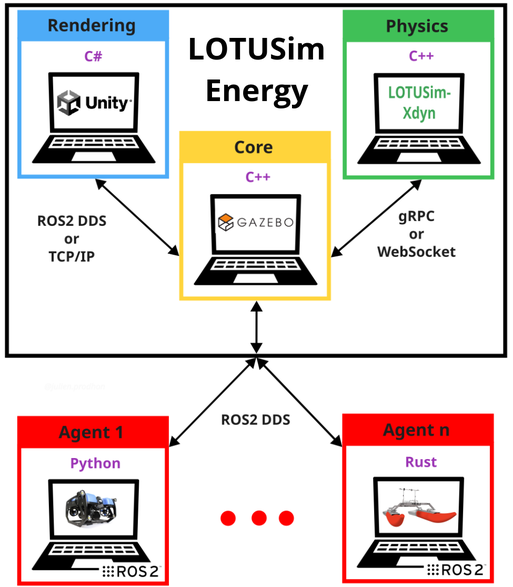}
    \caption{\emph{LOTUSim-Energy} architecture}
    \label{fig:lotusim}
\end{figure}

This architecture supports real-time human–robot interaction, scalable multi-vehicle simulation across air/surface/subsea domains, and accelerated-time runs for learning-based methods. This will allow users to perform complete maintenance on all the different parts of a wind turbine (blade, nacelle, monopile...) with manned, unmanned, or hybrid drones.
\emph{LOTUSim-Energy}’s server–client design allows computationally intensive tasks (e.g. hydrodynamics) to run remotely while Gazebo maintains a barrier-synchronised update loop. %
Figure~\ref{fig:lotusim} summarises the architecture\footnote{\scriptsize LOTUSim-Energy: \url{https://github.com/IRL-Crossing-CNRS/LOTUSim-Energy}}.

At each simulation timestep (user-configurable), Gazebo queries the three client classes via dedicated plugins: (i) \emph{Physics}, (ii) \emph{Rendering}, and (iii) \emph{Agent Interaction}. Each client computes updates, locally or remotely, and returns results before the next barrier, ensuring a consistent world state. To limit scheduling bias from asset load order, update requests are randomly permuted each tick (with a reproducible seed). Both real-time (Real Time Factor: RTF$\approx$1) and accelerated-time (RTF$>$1) modes are supported.

\subsection{Physics}
A unified physics interface exports each asset’s pose/twist, actuation, and environment parameters to external engines and re-ingests the resulting wrenches or updated poses. For marine vehicles, \emph{LOTUSim–Xdyn} includes USVs/AUVs/ROVs, with six-degree-of-freedom dynamics (including added mass, damping, wave, and current forces) connected via gRPC/WebSocket. Aerial agents use standard Gazebo rigid-body models and wind plugins. This abstraction allows users to swap or extend physics backends without changing the core simulator.

\subsection{Agent Interaction}
Agents interface through ROS~2 (DDS), exposing topics, actions, and services for guidance, planning, perception, and logging. This allows integration of existing autonomy stacks or real hardware bridges with minimal effort. For human–robot interaction, Unity provides operator UIs (desktop and virtual reality), supports authority handover, and logs operator actions for repeatable studies. By decoupling physics and visualisation from agent logic, \emph{LOTUSim–Energy} ensures consistent simulation timing while enabling flexible, real-time, or accelerated agent interaction.

\subsection{Rendering}
Rendering is optional and often disabled for large-scale training. When enabled, Unity subscribes to asset states and produces photorealistic views, along with synthetic sensor outputs (RGB/IR, depth, segmentation, lidar/sonar stubs) with pixel-accurate ground truth. Users can inject custom effects (e.g., spray, lighting, turbidity) via shaders to support perception research. These features will allow operators and learning algorithms to train more effectively, bridging the gap toward real-life deployments in wind farms.

\subsection{Human-in-the-Loop \& Multi-User}

\emph{LOTUSim-Energy} provides a high-fidelity Desktop/VR interface for collaborative training and mission oversight. Integrating Leap Motion and eye tracking, the platform supports gesture-based interaction and attention logging. Real-time synchronisation via Photon Unity Networking (PUN2) maintains a consistent world state for distributed teams through two specialised roles:
\begin{itemize}
    \item \textit{Embodied Operator:} Physics-constrained manual control for mission rehearsal.
    \item \textit{Supervisor/Spectator:} Free-flight mode for oversight and dynamic waypoint placement.
\end{itemize}

\section{LOTUSim-Energy Environment Model}
\label{sec:LOTUSim-environment-model}

\subsection{Surface}

\begin{figure}[H]
    \centering
    \includegraphics[width=0.45\linewidth]{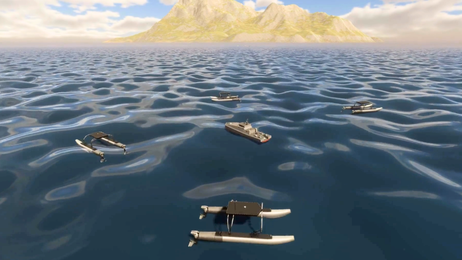}
    \caption{Surface agents deployed in \emph{LOTUSim-Energy}.}
    \label{fig:surface}
\end{figure}

In \emph{LOTUSim-Energy}, surface vessel dynamics are simulated using Xdyn\footnote{\scriptsize \url{https://gitlab.com/sirehna_naval_group/sirehna/xdyn}}, an open-source lightweight framework for real-time ship motion.
The solver implements Fossen’s equations of motion \cite{Fossen2011} and accounts for a wide range of hydrodynamic effects, including diffraction and Froude–Krylov forces.
In addition, Xdyn provides customisable manoeuvring models and actuator modules, with recent extensions incorporating wind-propulsion concepts for large cargo ships \cite{Babarit2024}.
For surface waves, Xdyn employs linear Airy wave theory \cite{Goda2010,Rodenbusch1986}, which models gravity-wave propagation under assumptions of uniform depth and constant fluid properties.
Although simplified, this formulation is computationally efficient and well-suited for simulating wave–hull interactions in real time.
Simulation outputs such as surface elevations and flow fields can be exported as 2D or 3D grids for further offline analysis.
In the forked version integrated into \emph{LOTUSim-Energy}, current profiles are also included, enabling a more complete treatment of the environmental disturbances acting on surface robots.
As a result, USVs and surface drones in our scenarios are directly influenced by realistic wave and wind forcing, a critical requirement for evaluating control and guidance strategies in offshore wind maintenance (see Fig.~\ref{fig:surface}).

\subsection{Aerial}
\begin{figure}[H]
    \centering
    \includegraphics[width=0.37\linewidth]{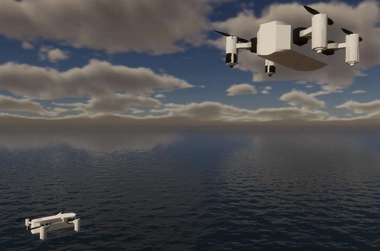}
    \includegraphics[width=0.25\linewidth]{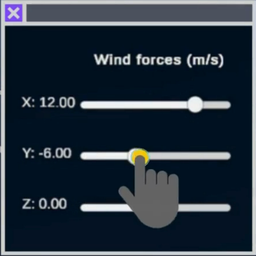}
    \caption{Left: X500 drone agents in \emph{LOTUSim-Energy}. Right: User wind interface to set different wind forces}
    \label{fig:air}
\end{figure}

Wind dynamics are handled through Gazebo’s wind plugin \cite{Koenig2004}, which supports dynamic wind fields varying in both time and space\footnote{\tiny\url{https://github.com/gazebosim/gz-sim/tree/gz-sim9/src/systems/wind_effects}}.
This allows users to configure realistic atmospheric conditions, ensuring that all aerial drones respond consistently to the imposed wind environment (Fig.~\ref{fig:air}).

\subsection{Underwater}

Realistic current modeling is critical for autonomous inspection near monopiles, cables, and subsea foundations, directly shaping vehicle guidance, energy-aware planning, and station-keeping. \emph{LOTUSim-Energy} implements an Ekman model \cite{Constantin2019} with three steady-state vertical layers. A surface layer, where wind stress and the Coriolis effect produce the Ekman spiral, and a bottom layer, shaped by frictional drag and bathymetric gradients, bound an intermediate geostrophic interior unaffected by either boundary. Together, these layers yield depth-dependent profiles relevant to robotic agents at different turbine zones.

We implemented two current models in Xdyn: a baseline constant flow and an Ekman-inspired formulation, where the velocity components $(u,v,w)$ are expressed as functions of $(x,y,z,t)$. This extended module is referred to as LOTUSim-Xdyn\footnote{\scriptsize LOTUSim-Xdyn is a modified version of Xdyn. Source Code: \url{https://github.com/naval-group/lxdyn}}.

\subsubsection{Domain and coordinate conventions}
We adopt a local ENU coordinate system tangent to WGS84 at the site origin, with $\Omega_h \subset \mathbb{R}^2$ the wind farm's horizontal footprint and $H : \Omega_h \to \mathbb{R}_{>0}$ the bathymetric depth (m). The space–time domain is:

\begin{multline}
\Omega = \left\{ (x, y, z, t) \;:\; (x, y) \in \Omega_h,\; z \in [-H(x, y),\, 0], \right.\\
\left. t \in [t_0, t_f] \right\}
\end{multline}

The current field $(u, v, w): \Omega \to \mathbb{R}^3$ gives velocity (m\,s$^{-1}$) along $(x,y,z)$; for Ekman-layer formulations we define:
\begin{align}
z_s(x, y, z) &= -z \quad \text{(depth from surface)} \\
z_b(x, y, z) &= H(x, y) + z \quad \text{(depth from seabed)}
\end{align}
with both \(z_s, z_b \in [0, H(x, y)]\).

\subsubsection{Parameters}
The current equations in each layer use the following parameters:
\begin{itemize}
    \item Coriolis parameter $f = 2 \Omega \sin(\varphi)$, with $\varphi$ the latitude and $\Omega$ Earth's rotation rate; its sign sets the spiral rotation (northern/southern hemisphere).
    \item Drag coefficient $C_D$, following Curcic and Haus \cite{Curcic2020}:
    \begin{equation}
        C_D =
        \begin{cases}
            (0.79 + 0.08\, U_{10}) \times 10^{-3}, & U_{10} < 20.5~\mathrm{m/s},\\[2pt]
            2.43 \times 10^{-3}, & U_{10} \ge 20.5~\mathrm{m/s}.
        \end{cases}
    \end{equation}
    \item Wind stress $\tau_s = C_D \rho_{\text{air}} U_{10}^{2}$ uses the $10$\,m wind $U_{10}$.
\end{itemize}

\subsubsection{Top layer}

Surface currents combine Airy wave orbital velocities with the Ekman spiral (northern hemisphere formulation):
\begin{equation} \label{eq: Article_Ekman_Top}
\begin{split}
&u = \bar{u} + V_0 e^{- \pi z_s / D_s} \cos\left(\frac{\pi}{4} - \frac{\pi z_s}{D_s} - \phi \right) \\
&v = \bar{v} + V_0 e^{- \pi z_s / D_s} \sin\left(\frac{\pi}{4} - \frac{\pi z_s}{D_s}  + \phi \right) \\
&w = 0
\end{split}
\end{equation}
where $z_s$ is the depth (null at the surface, positive downwards), $D_s$ is the surface Ekman layer depth, $V_0$ is the surface current velocity, and $\phi$ is the wind orientation at the surface.
This layer may be superimposed on Airy wave orbital velocities in the surface physics module

\subsubsection{Middle Layer}
\begin{equation} \label{eq: Article_Ekman_Middle}
u = \bar{u}, \quad v = \bar{v}, \quad w = 0
\end{equation}
The middle layer is unaffected by the waves or the seabed.

\subsubsection{Bottom Layer}

For irregular seabeds, additional correction terms are applied to maintain continuity and satisfy boundary conditions. In the northern hemisphere, this yields the following formulation of the bottom Ekman spiral:
\begin{equation} \label{eq: Article_Ekman_Bottom}
\scriptsize
\begin{split}
&u = \bar{u}\left[1 - e^{- \pi z_b / D_b}\cos\left(\frac{\pi z_b}{D_b}\right)\right] - \bar{v}e^{- \pi z_b / D_b}\sin\left(\frac{\pi z_b}{D_b}\right) \\
&v = \bar{u} e^{- \pi z_b / D_b}\sin\left(\frac{\pi z_b}{D_b}\right) + \bar{v}\left[1 - e^{- \pi z_b / D_b}\cos\left(\frac{\pi z_b}{D_b}\right)\right] \\
&w = 0
\end{split}
\end{equation}
where $z_b$ is the depth (null at the seabed, positive upwards) and $D_b$ is the bottom Ekman layer depth (with eddy viscosity $\nu_E$), governing the shear encountered by seabed-proximal tasks (e.g., scour and CP surveys).

\section{Autonomous offshore O\&M}
\label{sec:LOTUSim-Energy}

The following subsections describe how the environment, assets, vessel models, autonomy modules and operator interfaces are integrated to enable realistic operations planning, risk assessment, and rehearsal of common O\&M tasks.

\subsection{Environment configuration}

\noindent\emph{LOTUSim-Energy} ingests MetOcean products from the Copernicus Marine Service\footnote{\url{https://data.marine.copernicus.eu/products}} as boundary conditions and for day-specific calibration. Time-stamped wind, wave, and current fields from day \(D\) are assimilated to estimate Ekman-layer coefficients.
The calibrated model is then propagated to day \(D{+}1\) to generate physics-consistent forecasts of subsurface currents to support safety analysis, risk management, and scheduling of offshore wind O\&M operations.

\subsection{Scene}

\begin{figure}[H]
    \centering
    \includegraphics[width=0.29\linewidth]{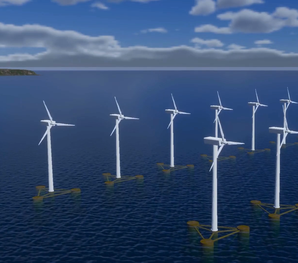} a
    \includegraphics[width=0.37\linewidth]{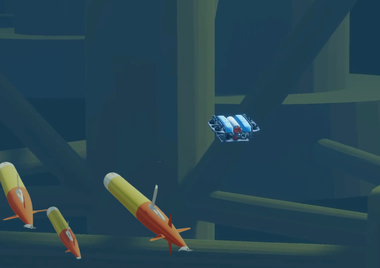} b
    \caption{Left: Illustration of a wind turbine farm scenario. Right: BlueROV and LRAUV operating underwater}
    \label{fig:agent-farm}
\end{figure}

\emph{LOTUSim-Energy} already ships with a pre-built Unity scene representing a multi-turbine offshore wind farm (Fig.~\ref{fig:agent-farm}a), so a wind-turbine maintenance scenario is immediately available. The scene includes turbine digital twins and environmental fields (wind, waves, currents) driven by \emph{LOTUSim-Xdyn}’s environment physics. It is wired to the simulator’s vehicle stack, allowing users to spawn UAV/USV/ROV agents and rehearse common O\&M primitives detailed in Table~\ref{tab:lotusim_tasks} (launch/recovery, waypoint transit, station-keeping, blade-face inspection passes, subsea transects) while logging time/energy budgets and safety hold states.

\subsection{Drones}

Subsea operations are critical for offshore O\&M, as most degradation, wear, and essential inspections occur underwater.
Currently, \emph{LOTUSim-Energy} supports two representative underwater agents (see Fig.~\ref{fig:agent-farm}b):
\begin{enumerate}
    \item a tethered BlueROV for close-range visual inspections, cathodic protection (CP) checks, and light intervention;
    \item a long-endurance LRAUV for wide-area surveys and cable or structure transects.
\end{enumerate}

Both vehicles operate in teleoperation, shared-control, or fully autonomous modes under the same metocean conditions described in Section~\ref{sec:LOTUSim-environment-model}.

In addition to these underwater drones, \emph{LOTUSim-Energy} provides X500 models for aerial operations and a WAMV vessel model for surface-level tasks.

\subsection{Battery Simulator Plugin}

To enable energy-aware mission planning, we developed a modular battery simulator plugin that couples state-of-charge estimation to each vehicle's instantaneous propulsive effort, rather than assuming a constant power draw. The plugin publishes real-time voltage and state-of-charge for use by mission planners and operators, enabling energy-aware evaluation of autonomy strategies under realistic power constraints.

\begin{table}[htbp]
    \caption{Representative offshore wind O\&M tasks}
    \label{tab:lotusim_tasks}

    \vspace{2pt}
    \parbox{.9\columnwidth}{\scriptsize \textbf{Agent type:} UAV (aerial), USV (surface), ROV (tethered subsea), AUV (untethered subsea), CTV (crew transfer vessel). \textbf{Sensors:} RGB (visible), IR (thermal), LiDAR (laser), MBES (multibeam), SSS (side-scan sonar), DVL (Doppler velocity log), INS (inertial nav.), UT (ultrasonic thickness), EO (electro-optical), AIS (Automatic Identification System), GPS. \textbf{Structures:} LE (leading edge),TP (transition piece), TLP (tension-leg platform). Rows shaded in blue are the tasks empirically demonstrated in Section~\ref{sec:experimental}; the remaining rows are supported by the underlying agent/sensor stack but have not yet been benchmarked end-to-end.}
    \vspace{5pt}

    \scriptsize
    \centering
    \begingroup
    \setlength{\tabcolsep}{2pt}
    \begin{tabular}{|>{\raggedright\arraybackslash}p{0.19\linewidth}|
                    >{\raggedright\arraybackslash}p{0.33\linewidth}|
                    >{\centering\arraybackslash}p{0.1\linewidth}|
                    >{\centering\arraybackslash}p{0.14\linewidth}|
                    >{\raggedright\arraybackslash}p{0.15\linewidth}|}
    \hline
    \textbf{Category} & \textbf{Task} & \textbf{Agent Type} & \textbf{Autonomy} & \textbf{Primary Sensors} \\
    \hline
    \cellcolor{blue!20}{Inspection (Topside)} & \cellcolor{blue!20}\makecell[l]{Blade surface scan\\(orbit pattern)} & \cellcolor{blue!20}UAV & \cellcolor{blue!20}\makecell[c]{Shared\\Auto} & \cellcolor{blue!20}\makecell[l]{RGB\\IR\\LiDAR} \\
    \hline
    \cellcolor{blue!20}Inspection (Topside) & \cellcolor{blue!20}\makecell[l]{Tower/TP corrosion\\\& fastener check} & \cellcolor{blue!20}UAV & \cellcolor{blue!20}\makecell[c]{Teleop\\Shared} & \cellcolor{blue!20}\makecell[l]{RGB\\LiDAR} \\
    \hline
    \cellcolor{blue!20}Inspection (Subsea) & \cellcolor{blue!20}\makecell[l]{Monopile/TP\\scour survey} & \cellcolor{blue!20}\makecell[c]{ROV\\AUV\\USV} & \cellcolor{blue!20}Auto & \cellcolor{blue!20}\makecell[l]{MBES\\SSS\\DVL\\INS} \\
    \hline
    Inspection (Subsea) & \makecell[l]{Cable route \&\\burial depth check} & \makecell[c]{AUV\\USV} & Auto & \makecell[l]{MBES\\SSS\\Magnetom.} \\
    \hline
    Inspection (Subsea) & \makecell[l]{Cathodic protection (CP)\\\& anode condition} & ROV & \makecell[c]{Teleop\\Shared} & \makecell[l]{CP probe\\Camera\\Sonar} \\
    \hline
    Inspection (Floating) & \makecell[l]{Mooring / TLP\\tendon inspection} & \makecell[c]{ROV\\AUV} & Auto & \makecell[l]{Sonar\\UT\\Camera} \\
    \hline
    Cleaning / Minor Repair & \makecell[l]{Biofouling removal\\(subsea)} & ROV & \makecell[c]{Teleop\\Shared} & \makecell[l]{Camera\\Sonar} \\
    \hline
    MetOcean Ops & \makecell[l]{Access-window\\rehearsal (CTV)} & \makecell[c]{USV\\CTV} & \makecell[c]{Auto\\Shared} & \makecell[l]{Wind\\AIS} \\
    \hline
    Environmental Monitoring & \makecell[l]{Marine-mammal\\exclusion zone (MMEZ)} & \makecell[c]{USV\\UAV} & Auto & \makecell[l]{Acoustics\\EO/IR\\Radar} \\
    \hline
    Emergency / Response & \makecell[l]{Man-overboard\\drill \& abort logic} & \makecell[c]{USV\\UAV} & \makecell[c]{Auto\\Shared} & \makecell[l]{AIS\\EO/IR\\GPS} \\
    \hline
    \end{tabular}
    \endgroup
\end{table}

\section{Experimental Inspection Scenario}
\label{sec:experimental}

\subsection{Simulation Environment and System Setup}

\begin{figure}[htbp!]
    \centering
    \includegraphics[width=0.60\linewidth]{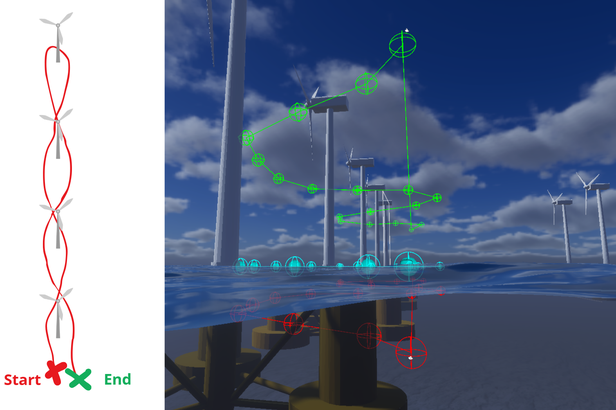}
    \caption{Cross-domain inspection path}
    \label{fig:inspection_path}
\end{figure}

As shown in Figure~\ref{fig:inspection_path}, a multi-domain inspection path for monopile and TP was defined, including surface, underwater, and aerial waypoint sequences (highlighted in blue in Table~\ref{tab:lotusim_tasks}):

\begin{itemize}
    \item Surface Monitoring: A support vessel or USV follows a wide-area trajectory to provide situational awareness and monitor the operational zone using an AIS-based waypoint follower.

    \item Underwater Inspection: The BlueROV2 performs a close-range survey of the submerged TP structural components. The trajectory enables high-resolution inspection to detect marine growth, corrosion, or structural defects such as cracks.

    \item Aerial Inspection: X500 drones conduct an aerial inspection of the monopile and above-water structures, enabling visual assessment of external surfaces and detection of corrosion or damage through onboard sensing and computer vision.
\end{itemize}

\subsection{Waypoint-Following Navigation Framework}

A Gazebo-integrated Waypoint Follower plugin guides vessels through 2D coordinate sequences via a closed-loop PID heading controller with bang--bang linear velocity regulation, initialised with conservative default values (zero heading/integral terms, vehicle at rest): a $0.5\,\mathrm{m}$ waypoint-reach tolerance, acceleration limits of $0.5\,\mathrm{m/s^2}$/$0.01\,\mathrm{rad/s^2}$ (linear/angular), and velocity saturation of $1$--$10\,\mathrm{m/s}$/$0.05\,\mathrm{rad/s}$. Dynamic replanning via a ROS2 service interface allows runtime adaptation to evolving mission objectives.

\subsection{AIS-Referenced Trajectory Following}

\begin{figure}[htbp]
    \centering
    \includegraphics[width=0.58\linewidth]{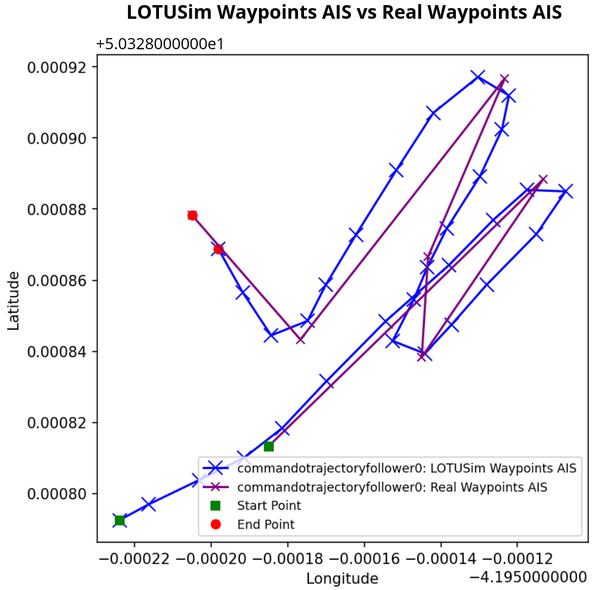}
    \caption{Real AIS plot compared to simulated plot for 10 points
    }
    \label{fig:ais_plot}
\end{figure}

The USV waypoint follower is driven along a waypoint sequence derived from a real AIS track of a small Class~B vessel (Fig.~\ref{fig:ais_plot}), illustrating AIS-referenced guidance rather than a controller-tracking benchmark, since the track reflects an independently controlled vessel with its own disturbances and GNSS noise, not a ground-truth reference.

\subsection{Visual Inspection Using Onboard Detection}

\begin{figure}[htbp]
    \centering

    \begin{subfigure}[b]{0.42\linewidth}
        \centering
        \includegraphics[width=\linewidth]{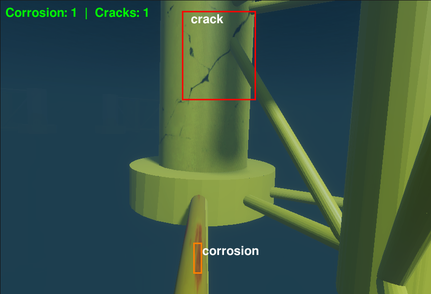}
    \end{subfigure}
    \hfill
    \begin{subfigure}[b]{0.42\linewidth}
        \centering
        \includegraphics[width=\linewidth]{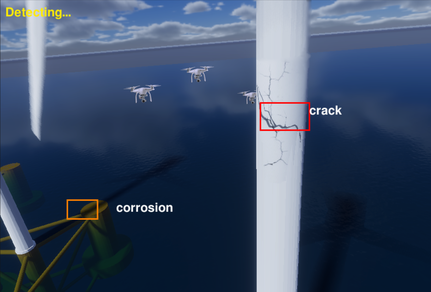}
    \end{subfigure}
    \caption{Fault detected (corrosion, cracks) by BlueROV (underwater) and X500 (air)}
    \label{fig:fault-detec}
\end{figure}

The platform integrates a YOLO-based real-time object detection framework\footnote{YOLO model: \url{https://tinyurl.com/yc3jr99t}} to identify structural anomalies such as cracks and corrosion. Figure~\ref{fig:fault-detec} illustrates detections performed underwater with a BlueROV2 and aerially with an X500 UAV during blade inspection.

\subsection{Deployment Use Cases \& Interaction}

\subsubsection{Ocean Current}

An actively propelled LRAUV commanded to hold a straight trajectory is exposed to the Ekman-based, vertically resolved current field. Despite active propulsion, depth-dependent lateral drift and rotational flow require continuous thruster compensation.

\subsubsection{HITL}

\begin{figure}[htbp]
    \centering
    \includegraphics[width=0.68\linewidth]{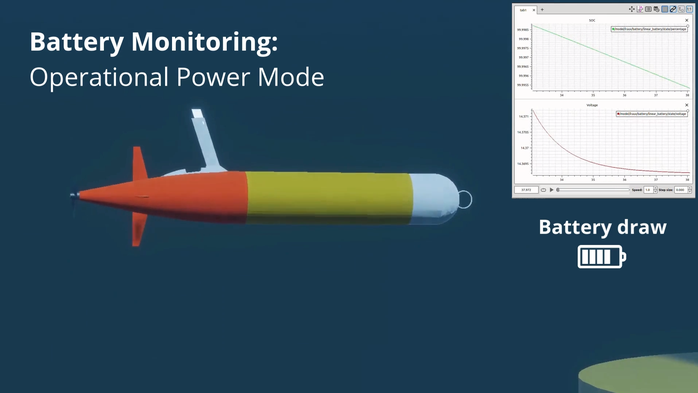}
    \caption{Real-time monitoring of LRAUV battery capacity and discharge via the plugin.}
\label{fig:lrauv_thruster}
\end{figure}

The operator interacts through a dual-layer interface: monitoring AIS data for vessels transiting to the mission zone, and, for inspection tasks, precision control of the onboard camera for structural assessment (e.g., cracks). This is augmented by the battery plugin, which tracks vehicle capacity and discharge (e.g., LRAUV in Fig.~\ref{fig:lrauv_thruster}), letting operators adapt inspection intensity to ensure reserve for safe recovery.

\begin{figure}[htbp]
    \centering
    \includegraphics[width=0.35\linewidth]{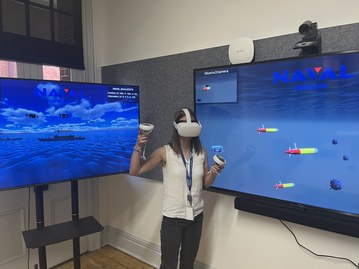}
    \hfill
    \includegraphics[width=0.46\linewidth]{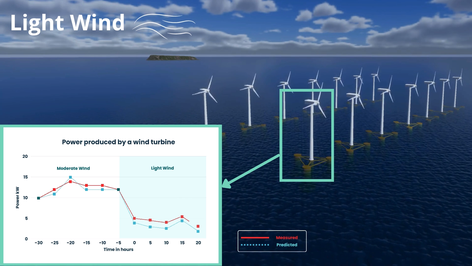} b
    \caption{Left: User interacting through VR. Right: Electricity production monitored  depending on wind strength}
    \label{fig:vr-wind-lotusim}
\end{figure}

Leap Motion further lets operators control the embarked camera (zooming, reorienting) via hand gestures, without a full VR headset. The same Desktop/VR interface (Fig.~\ref{fig:vr-wind-lotusim}) also drives an education mode with real-time wind/weather controls for stakeholder outreach.

\section{Conclusion}

We introduced \emph{LOTUSim-Energy}, a robotics-centric simulator unifying vehicle dynamics, environmental forcing, multi-user supervision, and benchmarking metrics for guidance, planning, and perception.
Future work will add physics-based underwater image formation for degraded-image datasets, a risk layer for biodiversity planning constraints, and benchmarking of energy-aware planners and fault detection under sim-to-real protocols.

\section*{Acknowledgements}

This work has been funded by the French Ministry for Europe and Foreign Affairs (MEAE) --- Franco-Australian Indo-Pacific Centre for Energy Transition (FACET).

\bibliographystyle{IEEEtran}
\bibliography{biblio}

\end{document}